\pdfoutput=1

\documentclass[11pt]{article}

\usepackage[preprint]{acl}

\usepackage{times}
\usepackage{latexsym}

\usepackage[T1]{fontenc}

\usepackage[utf8]{inputenc}

\usepackage{microtype}

\usepackage{inconsolata}

\usepackage{graphicx}
\usepackage{amsmath}
\usepackage{amsthm}
\usepackage{booktabs}
\usepackage{algorithm}
\usepackage{algorithmic}
\usepackage[switch]{lineno}
\usepackage{amsfonts, amssymb}
\usepackage{dsfont}
\usepackage{multirow}
\usepackage{makecell}
\usepackage{subfig}

\title{HUT: A More Computation Efficient Fine-Tuning Method With Hadamard Updated Transformation}

\author{
 \textbf{Geyuan Zhang\textsuperscript{1,2}},
 \textbf{Xiaofei Zhou\textsuperscript{1,2}\thanks{Corresponding Author.}},
 \textbf{Chuheng Chen\textsuperscript{1,2}}
\\
\\
 \textsuperscript{1}School of Cyber Security, University of Chinese Academy of Sciences,\\
 \textsuperscript{2}Institute of Information Engineering, Chinese Academy of Sciences,
\\
   {zhanggeyuan@iie.ac.cn, zhouxiaofei@iie.ac.cn, chenchuheng@iie.ac.cn}
}

\begin{document}
\maketitle
\begin{abstract}
Fine-tuning pre-trained language models for downstream tasks has achieved impressive results in NLP. However, fine-tuning all parameters becomes impractical due to the rapidly increasing size of model parameters. To address this, Parameter Efficient Fine-Tuning (PEFT) methods update only a subset of parameters. Most PEFT methods, such as LoRA, use incremental updates, which involve adding learned weight matrix increments to the original parameters. Although effective, these methods face limitations in capturing complex parameter dynamics and do not maintain a strong correlation between the original and updated parameters.
To overcome these challenges, we propose the direct \textbf{U}pdated \textbf{T}ransformation (UT) paradigm, which constructs a transformation directly from the original to the updated parameters. This approach ensures that the correlation between the original and updated parameters is preserved, leveraging the semantic features learned during pre-training.
Building on this paradigm, we present the \textbf{H}adamard \textbf{U}pdated \textbf{T}ransformation (HUT) method. HUT efficiently updates the original weight matrix using the Hadamard transformation with two low-rank matrices, offering a more expressive and flexible update mechanism. This allows HUT to capture richer parameter features through functional transformations, reducing computational complexity while maintaining or improving model quality.
Theoretical analysis and extensive experiments on RoBERTa and GPT-2 validate the effectiveness of HUT. Results show that HUT performs on par with or better than other PEFT methods in terms of model quality, while significantly reducing computational complexity.
\end{abstract}

\section{Introduction}

Pre-trained large language models have achieved great success in various natural language processing tasks. Typically trained on hyperscale corpora, these models are fine-tuned on downstream task datasets to improve performance. However, as the parameter size of these models increases, fine-tuning becomes computationally expensive.
Researchers have proposed two main lines of researches to handle this problem. One is In-Context Learning (ICL), which applies pre-trained models to downstream tasks without parameter adjustments by using prompt samples. However, wording \cite{DBLP:conf/naacl/WebsonP22} and ordering \cite{DBLP:conf/icml/ZhaoWFK021} in the prompt have a significant impact on the performance of the model, and studies \cite{DBLP:conf/nips/LianZFW22} have shown that ICL paradigms generally produce worse performance than fine-tuned paradigms.The other approach is Parameter Efficient Fine-Tuning (PEFT), which updates only a few parameters while keeping most fixed.

\begin{figure}[t]
\centering
\includegraphics[width=0.9\columnwidth]{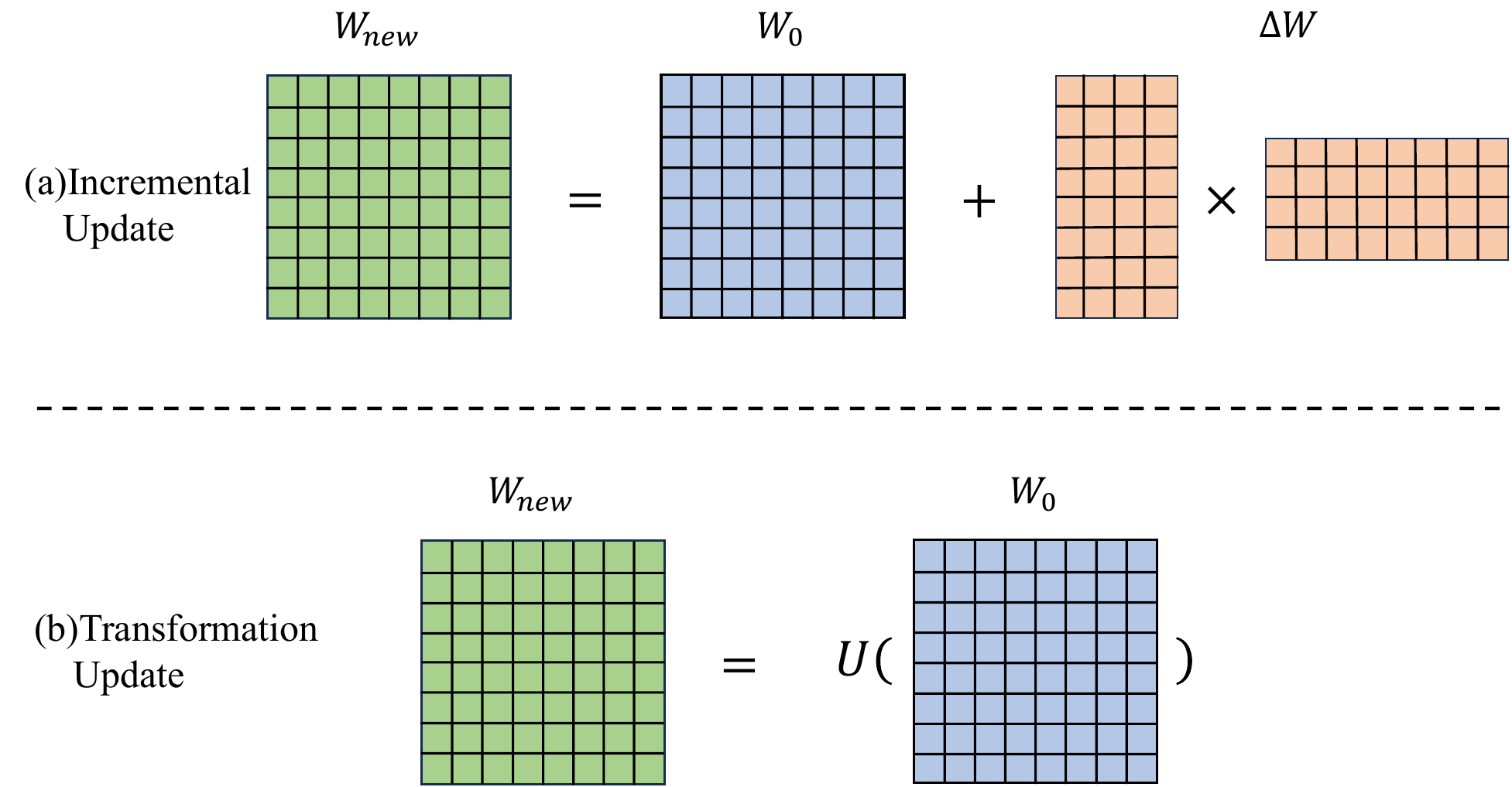} 
\caption{Parameter updating procedure through Incremental Update and our Transformation Update. Most of existing PEFT methods learn a incremental update by adding $\Delta W$ to original weight matrix $W_0$, while we proposed direct update method that uses an update transformation to get $W_{new}$.}
\vspace{-7mm}
\label{fig1}
\end{figure}

Existing PEFT methods typically update the weight matrix $W_0$ by incrementally adding $\Delta W$, as shown in Figure \ref{fig1}(a). These incremental update methods can be categorized into three groups.  The first group consists of Addition-based methods, such as Adapter \cite{DBLP:conf/icml/HoulsbyGJMLGAG19}, Prompt Tuning \cite{DBLP:conf/emnlp/LesterAC21}, and Prefix Tuning \cite{DBLP:conf/acl/LiL20}. These methods introduce a portion of the network layer or trainable parameters as $\Delta W$ to the original pre-trained model. During fine-tuning, only the added parameters are updated while the majority of other parameters are frozen. However, Adapter modifies the structure of the model, which can increase the inference latency. On the other hand, Prefix tuning and Prompt Tuning may make model optimization more challenging.
The second group is Specification-based methods, including BitFit \cite{DBLP:conf/acl/ZakenGR22}, Diffpuning \cite{DBLP:conf/acl/GuoRK20}, and FAR \cite{DBLP:conf/iscas/VuceticTZCMG22}. These methods directly specify certain parameters in the original model as $\Delta W$ to be trainable while keeping the remaining parameters frozen. Unlike Additive methods, Specification-based methods do not modify the original model structure. However, they are often not as effective. The last group of methods is Reparametrization-based methods, such as Lora \cite{DBLP:conf/iclr/HuSWALWWC22}, KronA \cite{DBLP:journals/corr/abs-2212-10650}, and AdaLoRA \cite{DBLP:conf/iclr/ZhangCBH0CZ23}. These methods reparameterize existing parameters into a parameter-efficient form. They decompose $\Delta W$ into a product of two or more low-rank matrices, which can be merged into the original weight parameters. As a result, there is no latency during the inference procedure.

Incremental update methods provide a straightforward and effective approach to training. However, they face significant limitations. Firstly, these methods do not maintain a strong correlation between the original parameters and the updated parameters. This lack of correlation means that the semantic information encoded in the pre-trained parameters is not fully leveraged during the fine-tuning process, potentially leading to suboptimal performance. Secondly, incremental update methods struggle to capture the complex dynamics of parameter interactions, as they primarily focus on linear updates. This linearity often fails to reflect the intricate changes needed for adapting large models to diverse downstream tasks. 

To address this issue and further enhance the performance of PEFT methods, we propose a direct Updated Transformation (UT) paradigm. In this paradigm, the original parameter is calculated directly using an updated transformation $U(\cdot)$ to obtain the updated parameter, denoted as $W_{new} = U(W_0)$. This ensures that the correlation between the original and updated parameters is preserved, leveraging the semantic features learned during pre-training. The UT paradigm is illustrated in Figure \ref{fig1} (b). Building upon this paradigm, we introduce a method called Hadamard Updated Transformation (HUT) for PEFT. HUT utilizes the Hadamard transformation, which consists of only two low-rank matrices, to update the original weight matrix. Compared to incremental methods, our HUT method not only significantly reduces computational complexity, but also captures richer parameter updated features through functional transformation. We conducted extensive experiments on a wide range of tasks and models to demonstrate the effectiveness of HUT. Specifically, we evaluated the performance of RoBERTa-large models on the natural language understanding (GLUE) task and GPT-2 on the natural language generation (E2E) dataset. Experimental results reveal that HUT performs on-par with or outperforms the baseline on most metrics, while maintaining similar or faster speeds with more participants than the baseline model. Thus, we can conclude that HUT effectively reduces computational complexity and improves performance on downstream tasks.

The contributions of our work are as follows:
\begin{itemize}
    \item We propose a direct \textbf{U}pdated \textbf{T}ransformation (UT) paradigm, a novel parameter updating paradigm, which enhances the ability to capture richer parameter features by maintaining a strong correlation between the original and updated parameters.  
    \item Upon the UT paradigm, we introduce the \textbf{H}adamard \textbf{U}pdated \textbf{T}ransformation (HUT). HUT uses the Hadamard transformation with two low-rank matrices, ensuring lower computational complexity and higher efficiency while capturing richer parameter features through a strong correlation between original and updated parameters.
    \item We evaluate HUT through extensive experiments on natural language understanding and generation tasks. Results show that HUT outperforms previous methods on most metrics, reducing computational complexity without increasing inference cost.
\end{itemize}

\begin{figure*}[t]
\vspace{-4mm}
\centering
\subfloat[\small Comparison of Inremental Update and UT Paradigm]{
                \includegraphics[width=0.57\textwidth]{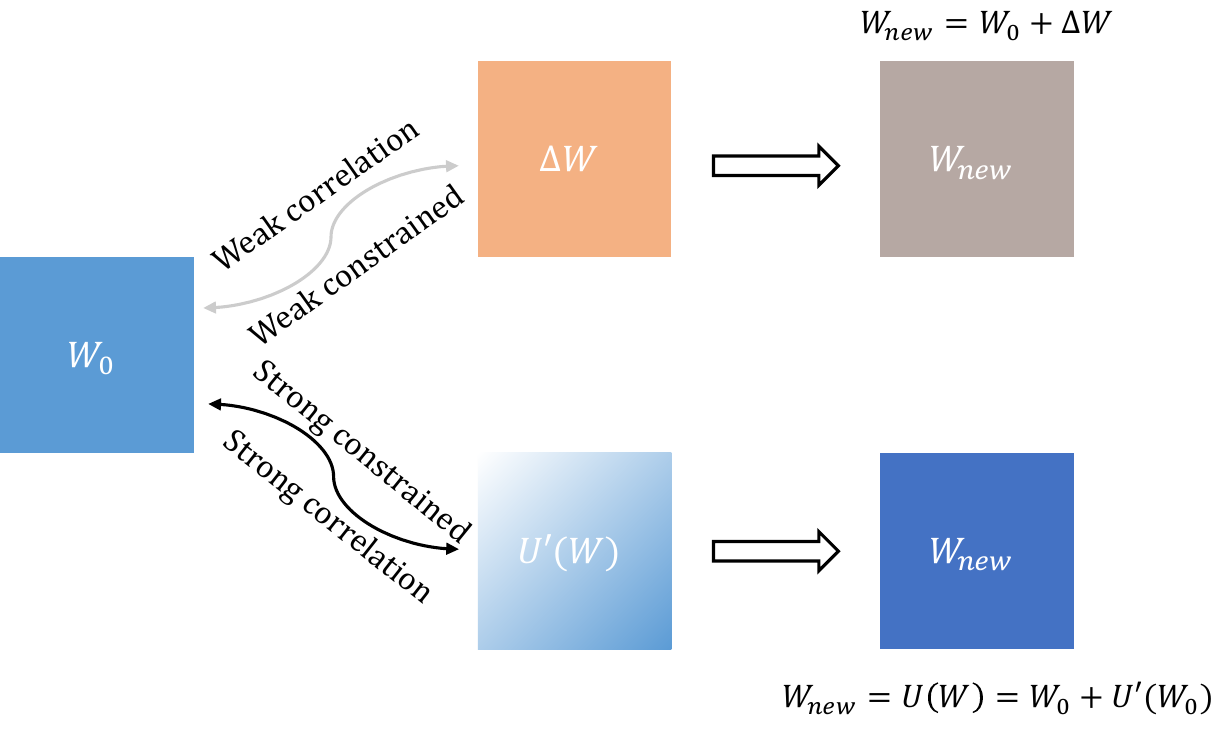}
                \label{fig2:a}}
\hfill
\subfloat[\small Architecture of HUT Module]{
                \includegraphics[width=0.36\textwidth]{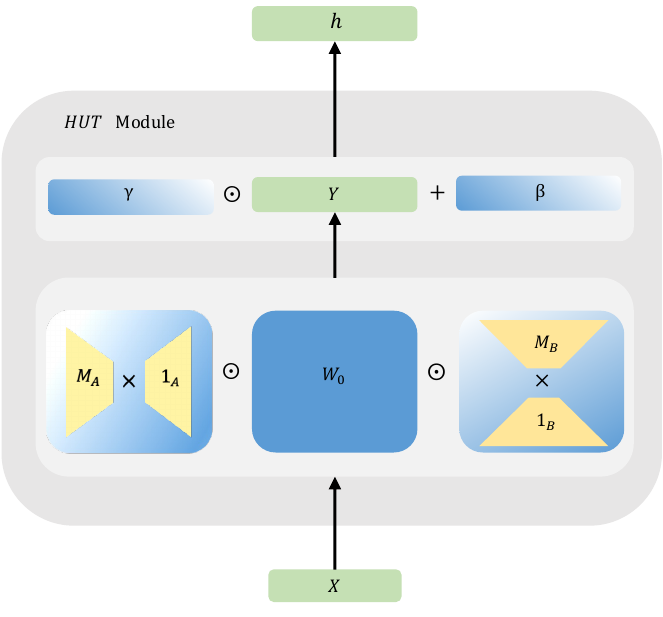}
                \label{fig2:b}}
\vspace{-1mm}
\caption{(a) Our proposed HUT can maintain a strong correlation between $W_0$ and $U'(W)$ so that the learned $U'(W)$ can leverage the semantic features learned during training. (b) The design of HUT Module. }
\label{fig2}
\vspace{-4mm}
\end{figure*}


\section{Related Work}

\textbf{Adapter} \cite{DBLP:conf/icml/HoulsbyGJMLGAG19}  approach inserts bottleneck shaped modules (called Adapter) into the Transformer layers. The Adapter layer first uses a down-sampling matrix $W_{down} \in \mathbb{R}^{d \times r}$ to project the input $x$ from a higher dimension $d$ to a smaller dimension $r$, followed by a nonlinear function $f(\cdot)$, and then uses an upsampling matrix $W_{up} \in \mathbb{R}^{r \times d}$ to transform it from dimension $r$ to $d$. The formulation is:
\begin{equation}
    \begin{aligned}
        h = x + f(xW_{down})W_{up}
    \end{aligned}
    \label{equ1}
\end{equation}

Houlsby et al. (2019) places two adapters sequentially within one layer of the transformer, one after the multi-head attention and one after the FFN sub-layer.

\textbf{LoRA} \cite{DBLP:conf/iclr/HuSWALWWC22} takes inspiration from Intrinsic SAID and hypothesize the updates to the weights also have a low “intrinsic rank” during adapting a large model to a specific downstream task. For a pre-trained model with weight $W_0 \in \mathbb{R}^{d \times k}$, the update of $W_0$ is decomposed into a product of two low-rank matrices. The forward process can be formulated as:
\begin{equation}
    \begin{aligned}
        h = x W_0 + x \Delta W = x W_0 + s \cdot x W_A W_B
    \end{aligned}
    \label{equ2}
\end{equation}
where $W_A \in \mathbb{R}^{d \times r}$, $W_B \in \mathbb{R}^{r \times k}$ and the rank $r \ll \min (d,k)$. $s \ge 1$ is a tunable scalar hyperparameter.In theory, LoRA can apply this update to all dense layer, but in original paper, it is applied to the query and value projection matrices in the multi-head attention sub-layer. 

\textbf{KronA} \cite{DBLP:journals/corr/abs-2212-10650} is similar with LoRA. While LoRA uses matrix product of two low-rank matrices to get the incremental update, KronA replaces the normal product by the Kronecker product: $\Delta W = W_A \otimes W_B$. Kronecker product is not rank deficient, so it can maintains the rank of the input matrix. Similar to LoRA, KronA has a fixed scale factor, $s$, which is a hyperparameter.
\begin{equation}
    \begin{aligned}
        h = x W_0 + s \cdot x (W_A \otimes W_B)
    \end{aligned}
    \label{equ4}
\end{equation}

\textbf{AdaLoRA} \cite{DBLP:conf/iclr/ZhangCBH0CZ23} is a variant of LoRA. LoRA need to pre-specify the rank $r$ of each incremental matrix $\Delta W$ identical, leading to ignore the fact that the importance of weight matrices varies significantly across modules and layers when fine-tuning pre-trained models. So AdaLoRA is proposed, which dynamically allocates the parameter budget among weight matrices during LoRA-alike finetuning. AdaLoRA uses SVD-based adaptation to formulate the incremental matrices in the form of singular value decomposition:
\begin{equation}
    \begin{aligned}
        W = W_0 + \Delta W = W_0 + P \Lambda Q
    \end{aligned}
    \label{equ5}
\end{equation}
where $P \in \mathbb{R}^{d \times r}$ and $Q \in \mathbb{R}^{r \times k}$ represent the left/right singular vectors of $\Delta W$ and the diagonal matrix $\Lambda \in \mathbb{R}^{r \times r}$ contains the singular values $\{ \lambda_i \}_{1 \le i \le r}$ with $r \ll \min (d,k)$. The SVD-based adaptation is applied to every weight matrix of each transformer layer. 
In order to control budget, AdaLoRA proposes importance-aware rank allocation, which prunes redundant singular values based on a newly-designed importance metric. To be specific, in the process of incremental update with the form of singular value decomposition, the unimportant singular values are pruned according to the importance metric, so as to assign higher rank to the incremental matrix with high importance score.

\section{Method}

We describe the UT paradigm and the simple design of HUT. 
The principles outlined here apply to any dense layer in a deep learning model, although we only focused on certain weights in the Transformer language model as incentive use cases in our experiments. 

\subsection{Direct Updated Transformation (UT) paradigm}
Assuming the new learned parameters are represented by $W_{new}$, and the initial parameters are represented by $W_0$. For other PEFT methods that use incremental updates, such as LoRA \cite{DBLP:conf/iclr/HuSWALWWC22}, $W_{new} = W_0 + \Delta W$ where $\Delta W = sAB$. It can be seen that this $\Delta W$ has no much correlation with the original parameters $W_0$ and it is not constrained by the value of $W_0$. We believe that this results in the semantic information encoded in $W_0$ not being fully leveraged during the fine-tuning process. To address this limitation, we propose a new paradigm, the direct UT paradigm, which uses a updated transformation $U( \cdot )$ to directly update $W_0$. The formulation is:
\begin{equation}
     \begin{aligned}
         W_{new} = U(W_0)
     \end{aligned}
     \label{equ6}
 \end{equation}

For comparison, $W_{new}$ can be further expressed as:
\begin{equation}
         W_{new} = W_0 + U'(W_0)
     \label{equ10}
 \end{equation}
Therefore we can let $\Delta W = U'(W_0)$. This formulation ensures a strong correlation between $\Delta W$ and $W_0$, maintaining the semantic features learned during pre-training. We believe that preserving this relevance and constraint is crucial during the fine-tuning stage, as it allows the model to better utilize the pre-trained semantic information encoded in $W_0$.

\subsection{Hadamard Updated Transformation (HUT)}
\textbf{Hadamard Product.} The most intuitive form of UT is to apply a linear transformation to the weight matrix $W_0 \in \mathbb{R}^{d \times k}$ using a transformation matrix: $U(W_0)=T \times W_0$, where $T \in \mathbb{R}^{d \times d}$ is a transformation and $\times$ is matrix multiplication. But as mentioned before, matrix multiplication has high computation complexity, so we can use the Hadamard Product to implement the transformation:
\begin{equation}
    \begin{aligned}
        A \odot B = \left[\begin{array}{cccc}a_{11} b_{11} & a_{12} b_{12} & \cdots & a_{1 k} b_{1 k} \\ a_{21} b_{21} & a_{22} b_{22} & \cdots & a_{2 k} b_{2 k} \\ \vdots & \vdots & & \vdots \\ a_{d 1} b_{d 1} & a_{d 2} b_{d 2} & \cdots & a_{d k} b_{d k}\end{array}\right]
    \end{aligned}
    \label{equ7}
\end{equation}
where $A \in \mathbb{R}^{d \times k}$, $B \in \mathbb{R}^{d \times k}$ and $\odot$ indicates Hadamard product. From Eq.(\ref{equ7}), we can find that the Hadamard product operation requires that both matrices have the same shape. Contrast with matrix multiplication, Hadamard product has lower computation complexity. So we can use $\odot$ instead of $\times$ to reduce the computation complexity of the transformation. 

\textbf{Design of HUT.} Intrinsic SAID \cite{DBLP:conf/acl/AghajanyanGZ20} finds that the pre-trained language models have a low "instrisic dimension" and can still learn efficiently despite a random projection to a smaller subspace. So we hypothesize that the linear tranformation of the weight matrices through Hadamard product have a low "instrisic dimension" during updating procedure in Eq.(\ref{equ6}) and the transformation matrix $T$ is also have a low "intrisic rank". Further, to improve the representation ability of the tranformation,  we use two low-rank transformation matrices $M_A \in \mathbb{R}^{d \times r}$ and $M_B \in \mathbb{R}^{r \times k}$, where the rank $r \ll \min(d,k)$. The new updated transformation can be formulated by:
\begin{equation}
    \begin{aligned}
        W_{new} = \frac{(M_A \times \mathds{1}_A)}{r}  \odot W_0 \odot \frac{(\mathds{1}_B \times M_B)}{r}
    \end{aligned}
    \label{equ8}
\end{equation}
Where $\mathds{1}_A \in \mathbb{R}^{r \times k}$, $\mathds{1}_B \in \mathbb{R}^{d \times r}$, and they are used to map the shape of $M_A$ and $M_B$ to be the same with $W_0$. We call the parameter update form shown in Eq.(\ref{equ8}) as the Hadamard Transformation Updated transformation(HUT). While in code implementation, the $\times$ operation can be replaced, and we will discuss it later. 

In addition, according to \cite{DBLP:conf/nips/LianZFW22}, scaling and shifting the deep features can improve the performance of fine-tuning. Therefore, for the forward process $h = W_0 x$, we add scaling and shifting to the input features and update the parameter with meta weights, our modified forward pass yields:
\begin{equation}
    \begin{aligned}
        h = \gamma \odot (x \times W_{new}) + \beta
    \end{aligned}
    \label{equ9}
\end{equation}
Where $\gamma \in \mathbb{R}^{1 \times k}$ and $\beta \in \mathbb{R}^{1 \times k}$ are the scale and shift factors. Though the shape of $\beta$ and $\gamma$ are not same with $x$, we can use broadcasting notation\cite{DBLP:journals/cse/WaltCV11} to automatic expand their dimension during calculating.

\textbf{Apply HUT.} There are many weight matrices in the Transformer architecture, including $W_q, W_k, W_v, W_o$ in self-attention module and $W_d, W_u$ in FFN module. In principle, we can apply HUT to any subset of weight matrices mentioned above. 

\subsection{Computation Complexity Analysis}
We compare the most widely used PEFT method LoRA with HUT in Floating Points Operations(FLOPs). Suppose that the input $x \in \mathbb{R}^{N \times d}$ and the weight matrix $W_0 \in \mathbb{R}^{d \times k}$. Before the compare, we convert the Eq.(\ref{equ9}) into the following form:
\begin{equation}
    \begin{aligned}
        h = x \times (\gamma \odot m_A &m_B \odot W_0) + \beta, \\
        m_A = \frac{1}{r} \sum_{j=1}^{r} {M_A}_{i,j}, \ \ &m_B = \frac{1}{r} \sum_{i=1}^{r} {M_B}_{i,j}
    \end{aligned}
    \label{equ10}
\end{equation}
Where $m_A \in \mathbb{R}^{d \times 1}$ and $m_B \in \mathbb{R}^{1 \times k}$. Then we can compute the FLOPs of HUT in one forward process according to Eq.(\ref{equ10}), which is $(2d-1)Nk+4dk+rd+rk$. And according Eq.(\ref{equ2}), the FLOPs of one forward pass of LoRA is $(2d-1)Nk + (2r+1)dk$. For simplicity, let us assume that $d=k$, then the $\Delta \text{FLOPs}$ of LoRA and HUT in one forward process is:
\begin{equation}
    \begin{aligned}
        \Delta \text{FLOPs} &= \text{FLOPs}_{LoRA}-\text{FLOPs}_{HUT} \\
        &=2rd^2-3d^2-2rd
    \end{aligned}
    \label{euq11}
\end{equation}
Since $r \ll d$, we can ignore the last item $2rd$, and then we can get that $\Delta \text{FLOPs}=2rd^2-3d^2$. As a result, in theory, when $r \ge 2$, the FLOPs of mew is smaller than Lora. While LoRA is usually used with $r=4$ or $r=8$ empirically, in these cases, using HUT instead of LoRA can reduce the number of FLOPs. 

Moreover, during inference process, we can re-parameterize meta weights into the previous linear layer as the form of Eq.(\ref{equ10}), so HUT do not introduce any additional inference latency to original model. 

\section{Experiments}

\begin{table*}[t!]
  \centering
  \footnotesize
  \addtolength{\tabcolsep}{-1pt}
  \begin{tabular}{l|r|ccccccc}
  \hline
  \toprule
  Model \& Method & \# Trainable & \multicolumn{7}{c}{} \\
         & Parameters & SST-2 & MRPC & CoLA & QNLI & RTE & STS-B & Avg. \\
  \midrule
  $\text{RoB}_\text{large}$ (FT)* & 355.0M & 96.4 & 90.9 & 68.0 & 94.7 & 86.6 & 92.4 & 88.2 \\
  \midrule
  $\text{RoB}_\text{large}$ (PAdapter)* & 3.0M & 96.1\textsubscript{$\pm$.3} & 90.2\textsubscript{$\pm$.7} & 68.3\textsubscript{$\pm$1.0} & 94.8\textsubscript{$\pm$.2} & 83.8\textsubscript{$\pm$2.9} & 92.1\textsubscript{$\pm$.7} & 87.6 \\
  $\text{RoB}_\text{large}$ (PAdapter)* & 0.8M & \textbf{96.6}\textsubscript{$\pm$.2} & 89.7\textsubscript{$\pm$1.2} & 67.8\textsubscript{$\pm$2.5} & 94.8\textsubscript{$\pm$.3} &  80.1\textsubscript{$\pm$2.9} & 91.9\textsubscript{$\pm$.4} & 86.8 \\
  $\text{RoB}_\text{large}$ (HAdapter)* & 6.0M & 96.2\textsubscript{$\pm$.3} & 88.7\textsubscript{$\pm$2.9} & 66.5\textsubscript{$\pm$4.4} & 94.7\textsubscript{$\pm$.2} &  83.4\textsubscript{$\pm$1.1} & 91.0\textsubscript{$\pm$1.7} & 86.8 \\
  $\text{RoB}_\text{large}$ (HAdapter)* & 0.8M & 96.3\textsubscript{$\pm$.5} & 87.7\textsubscript{$\pm$1.7} & 66.3\textsubscript{$\pm$2.0} & 94.7\textsubscript{$\pm$.2} & 91.5\textsubscript{$\pm$.1} & 72.9\textsubscript{$\pm$2.9} & 84.9 \\
  $\text{RoB}_\text{large}$ (LoRA)* & 0.8M & 96.2\textsubscript{$\pm$.5} & 90.2\textsubscript{$\pm$1.0} & 68.2\textsubscript{$\pm$1.9} & \textbf{94.8}\textsubscript{$\pm$.3} & 85.2\textsubscript{$\pm$1.1} & 92.3\textsubscript{$\pm$.5} & 87.8 \\
  $\text{RoB}_\text{large}$ (VeRA)* & 0.061M & 96.1\textsubscript{$\pm$.1} & 90.9\textsubscript{$\pm$.7} & 68.0\textsubscript{$\pm$.8} & 94.4\textsubscript{$\pm$.2} & 85.9\textsubscript{$\pm$.7} & 91.7\textsubscript{$\pm$.8} & 87.8 \\
  $\text{RoB}_\text{large}$ (FourierFT)* & 0.048M & 96.0\textsubscript{$\pm$.5} & 90.9\textsubscript{$\pm$.3} & 67.1\textsubscript{$\pm$1.4} & 94.4\textsubscript{$\pm$.4} & 87.4\textsubscript{$\pm$1.6} & 91.9\textsubscript{$\pm$.4} & 88.0 \\
  $\text{RoB}_\text{large}$ (HUT) &0.9M & 96.1\textsubscript{$\pm$.1} & \textbf{91.0}\textsubscript{$\pm$.2} & \textbf{70.5}\textsubscript{$\pm$1.2} & 94.2\textsubscript{$\pm$.1} & \textbf{87.4}\textsubscript{$\pm$0.3} & \textbf{92.3}\textsubscript{$\pm$.1} & \textbf{88.6} \\
  \bottomrule
  \end{tabular}
  \caption{Results with RoBERTa-large on GLUE development set. The best results on each dataset are shown in bold. We report Matthew’s correlation for CoLA, Pearson correlation for STS-B, and accuracy for other tasks. Higher is better for all metrics. * indicates numbers published in prior works.
  }
  \vspace{-3mm}
  \label{tab:NLU_results}
\end{table*}

\subsection{Experimental Settings}

We evaluate the downstream task performance of HUT on RoBERTa-large \cite{DBLP:journals/corr/abs-1907-11692} and GPT-2 \cite{Radford2019LanguageMA}. Our experiments cover a wide range of tasks, from natural language understanding (NLU) to generation (NLG). Specifically, we evaluate on the GLUE \cite{DBLP:conf/iclr/WangSMHLB19} benchmark for RoBERTa. We follow the setup of \cite{DBLP:conf/acl/LiL20} on GPT-2 for a direct comparison. We use NVIDIA RTX3090 for all experiments.

We compare our methods with these types of approaches as follows: full fine-tuning (FT), BitFit \cite{DBLP:conf/acl/ZakenGR22}, HAdapter \cite{DBLP:conf/icml/HoulsbyGJMLGAG19}, LAdapter \cite{DBLP:conf/emnlp/LinMF20}, PAdapter \cite{DBLP:conf/eacl/PfeifferKRCG21}, LoRA \cite{DBLP:conf/iclr/HuSWALWWC22}, VeRA \cite{vera} and FourierFT \cite{fourierft}. See more details in Appendix \ref{sec:baselines}.

\begin{table}[]
    \centering
    \begin{tabular}{l|r|c}
    \hline
    \toprule
    Model \& Method & \# Trainable & FLOPs\\
         & Parameters  & (GFLOPs) \\
    \midrule
    $\text{RoB}_\text{large}$ (PAdapter) & 3.0M & 2.40 \\
    $\text{RoB}_\text{large}$ (PAdapter) & 0.8M & 1.80 \\
    $\text{RoB}_\text{large}$ (HAdapter) & 6.0M & 1.61 \\
    $\text{RoB}_\text{large}$ (HAdapter) & 0.8M & 0.41 \\
    $\text{RoB}_\text{large}$ (LoRA) & 0.8M & 0.86 \\
    $\text{RoB}_\text{large}$ (HUT) & 0.9M & \textbf{0.20} \\
    \bottomrule
    \end{tabular}
    \caption{We compare our HUT with other baselines in FLOPs based on NLU tasks as mentioned before.}
    \vspace{-5mm}
    \label{tab:flops}
\end{table}

\subsection{Natural Language Understanding}

\begin{figure}[t]
\centering
\includegraphics[width=0.9\columnwidth]{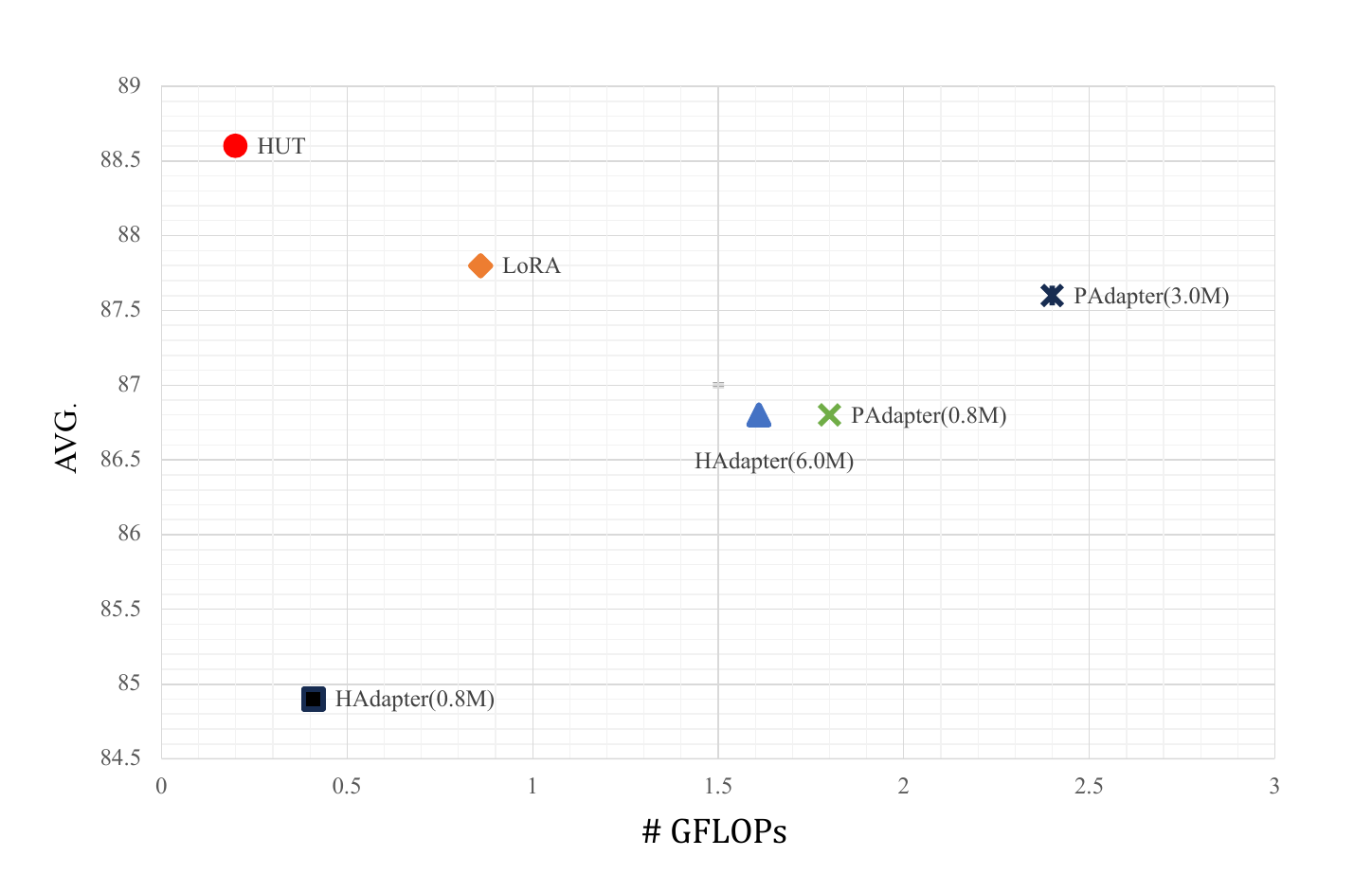} 
\caption{Average scores in GLUE benchmark based on RoBERTa with different PEFT methods. The x-axis is the number of GFLOPs, which indicates the computation complexity, and the y-axis is the average scores.}
\vspace{-3mm}
\label{fig3}
\end{figure}

\begin{table*}[t]
\small
\centering
\begin{tabular}{l|r|ccccc}
\hline
\toprule
Model \& Method & \# Trainable & \multicolumn{5}{c}{E2E NLG Challenge} \\
       & Parameters & BLEU & NIST & MET & ROUGE-L & CIDEr \\
\midrule
GPT-2 M (FT)* & 354.92M                         & 68.2 &	8.62 &	46.2 &	71.0 &	2.47  \\
GPT-2 M (LAdapter)* & 0.37M  & 66.3 &	8.41 &	45.0 &	69.8 &	2.40  \\
GPT-2 M (LAdapter)* & 11.09M & 68.9 &	8.71 &	46.1 &	71.3 &	2.47  \\
GPT-2 M (HAdapter)* & 11.09M & 67.3\textsubscript{$\pm$.6} & 8.50\textsubscript{$\pm$.07}	& 46.0\textsubscript{$\pm$.2} & 70.7\textsubscript{$\pm$.2}	& 2.44\textsubscript{$\pm$.01}        \\
GPT-2 M ($\text{FT}^{\text{Top2}}$)*   & 25.19M & 68.1 & 8.59 & 46.0  &  70.8 & 2.41  \\
GPT-2 M (PreLayer)* & 0.35M & 69.7 & 8.81 & 46.1 & 71.4 & 2.49  \\
GPT-2 M (LoRA)* & 0.35M & 70.4\textsubscript{$\pm$.1} & 8.85\textsubscript{$\pm$.02} & 46.8\textsubscript{$\pm$.2} & 71.8\textsubscript{$\pm$.1} & 2.53\textsubscript{$\pm$.02} \\
GPT-2 M (VeRA)* & 0.098M & 70.1 & 8.81 & 46.6 & 71.5 & 2.50 \\
GPT-2 M (FourierFT)* & 0.048M & 69.1\textsubscript{$\pm$.1} & 8.82\textsubscript{$\pm$.05} & \textbf{47.0\textsubscript{$\pm$.3}} & 71.8\textsubscript{$\pm$.1} & 2.51\textsubscript{$\pm$.02} \\
GPT-2 M (HUT) & 0.45M & \textbf{70.4}\textsubscript{$\pm$.1} & \textbf{8.86}\textsubscript{$\pm$.02} & 46.7\textsubscript{$\pm$.2} & \textbf{72.1\textsubscript{$\pm$.1}} & \textbf{2.54\textsubscript{$\pm$.01}} \\
\bottomrule
\end{tabular}
\caption{GPT-2 medium (M) with different adaptation methods on the E2E NLG Challenge. For all metrics, higher is better. Confidence intervals are shown for experiments we ran. * indicates numbers published in prior works.
}
\label{tab:gpt2_ft_results}
\end{table*}

\subsubsection{Models and Datasets.} We use GLUE \cite{DBLP:conf/iclr/WangSMHLB19} benchmark to evaluate the performence of our methds based on RoBERTa-large \cite{DBLP:journals/corr/abs-1907-11692} model in natural language understanding tasks. GLUE benchmark is a wide-ranging collection of natural language understanding tasks. Dataset details are summarized in Appendix \ref{sec:datasets}. RoBERTa-large \cite{DBLP:journals/corr/abs-1907-11692} consists of 357 millions parameters, and we take the pre-trained RoBERTa-large from HuggingFace Transformers library \cite{DBLP:conf/emnlp/WolfDSCDMCRLFDS20}.

\subsubsection{Implementation Details.} We apply HUT to query and value matrices \{$W_q$, $W_v$\} in self-attention module and set $r=8$, using AdamW \cite{DBLP:conf/iclr/LoshchilovH19} optimizer to train it for all sub-tasks. For HAdapter \cite{DBLP:conf/icml/HoulsbyGJMLGAG19}, PAdapter \cite{DBLP:conf/eacl/PfeifferKRCG21} and LoRA \cite{DBLP:conf/iclr/HuSWALWWC22}, we follow the original setup introduced in \cite{DBLP:conf/iclr/HuSWALWWC22}. While the other PEFT methods use a pre-trained model which is already adapted to MNLI to initialize the model for MRPC, RTE, and STS-B, we start with the original pre-trained RoBERTa large model. See Appendix for details on the hyperparameters used. We report the Matthew’s correlation for CoLA, Pearson correlation for STS-B, and accuracy for other tasks. More details please refer to Appendix \ref{sec:experiments}.

\subsubsection{Main Results.} Table \ref{tab:NLU_results} shows the experimental results on the GLUE validation dataset. We report mean of 5 runs using different random seeds. We can see that on four of the six datasets of the GLUE benchmark, we achieve the SOTA performance, and we achieve the best average score on all datasets. The datasets that achieve SOTA results are MRPC, CoLA, RTE and STS-B. On CoLA dataset, we achieve a significant performance improvement compared to the previous SOTA model LoRA, with an improvement of 2.3\% respectively. On the average score of all six datasets, the improvement is 0.6\% compared with previous SOTA FourierFT. 


Further more, we conduct experiments to compare the computation complexity between different methods mentioned in this section using FLOPs. The results are shown in table \ref{tab:flops} and figure \ref{fig3}. You can find that though HUT has more parameter than some basline methods, there are much fewer FLOPs than them during training and inference procedure and not introducing any inference latency. Figure \ref{fig3} visually shows the relationship between FLOPs and GLUE benchmark average scores of different models. Our proposed method has the least FLOPs and the highest GLUE benchmark average score. This indicates that HUT can not only reduce the computation complexity but also improve the model preformance. We believe that this is due to the fact that our proposed Hadamard updated transformation method can capture richer parameter updated features with efficient computation.

\subsection{Natural Language Generation}
\subsubsection{Models and Datasets.} HUT has been shown that it can get competitive results compared with other PEFT methods and full fine-tuning on NLU tasks, and we hope to answer if HUT still prevails on NLG tasks. So we use E2E NLG Challenge to evaluate the performence of our methods. It is first introduced in \cite{DBLP:conf/sigdial/NovikovaDR17} as a dataset for training end-to-end, data-driven natural language generation systems and is commonly used for data-to-text evaluation. And we use GPT-2 media as our base model which consists of over 354 millions parameters. We use the official evaluation script, which reports BLEU\cite{DBLP:conf/acl/PapineniRWZ02}, NIST\cite{DBLP:conf/eacl/BelzR06}, METEOR\cite{DBLP:conf/wmt/LavieA07}, ROUGE-L\cite{lin-2004-rouge}, and CIDEr\cite{DBLP:conf/cvpr/VedantamZP15}.

\begin{table*}[h]
\small
  \centering
  \begin{tabular}{l|cccccccc}
  \hline
  \toprule
                & \multicolumn{8}{c}{\# of Trainable Parameters = 0.8M} \\
  \midrule
  Weight Type           & $W_q$  & $W_k$  & $W_v$  & $W_o$      & $W_q,W_k$     & $W_q,W_v$     &   $W_q, W_k, W_v$   & $W_q, W_k, W_v, W_o$      \\
  Rank $r$              & 16      &  16     &  16     &   16        &   8           &   8         &          4          &           2                       \\
  \midrule
  MRPC  & 88.2   & 87.7   & 90.2   & 90.9       & 85.3     &  91.2      & 88.7  & 70.8             \\
  CoLA  & 61.7   & 60.7   & 66.8   & 68.2       & 60.4     & 71.7       &64.5   & 64.6              \\
  \bottomrule
  \end{tabular}
  \caption{Validation accuracy on MRPC and CoLA after applying HUT to different types of attention weights in RoBERTa-large, given the approximate number of trainable parameters.}
  \vspace{-3mm}
  \label{tab:weight_type}
\end{table*}

\begin{table*}[h]
\small
  \centering
  \begin{tabular}{r|c|ccccc}
  \hline
  \toprule
                                        & Weight Type           & $r=1$  & $r=2$  & $r=4$  & $r=8$  & $r=64$  \\
  \midrule
  \multirow{3}{*}{MRPC}  & $W_{v}$               & 90.2   & 90.9   & 90.0   & 90.4   & 89.7    \\
                         & $W_q, W_v$            & 90.7   & 88.5   & 88.7   & 91.2   & 90.9    \\
                         & $W_q, W_k, W_v, W_o$  & 68.4   & 70.8   & 71.6   & 68.4   & 70.8    \\
  \midrule
  \multirow{3}{*}{CoLA}& $W_v$                 & 67.5   & 66.9   & 69.5   & 70.0   & 66.8    \\ 
                         & $W_q, W_v$          & 69.5   & 67.1   & 68.4   & 71.7   & 68.1    \\ 
                         & $W_q, W_k, W_v, W_o$  & 54.2   & 64.6   & 64.6   & 49.6   & 43.1    \\
  
  \bottomrule
  \end{tabular}
  \caption{Validation accuracy on MRPC and CoLA with different rank $r$.}
  \vspace{-5mm}
  \label{tab:effect_r}
\end{table*}

\begin{figure*}[htbp]
\centering
\subfloat[HUT]{
                \includegraphics[width=0.47\textwidth]{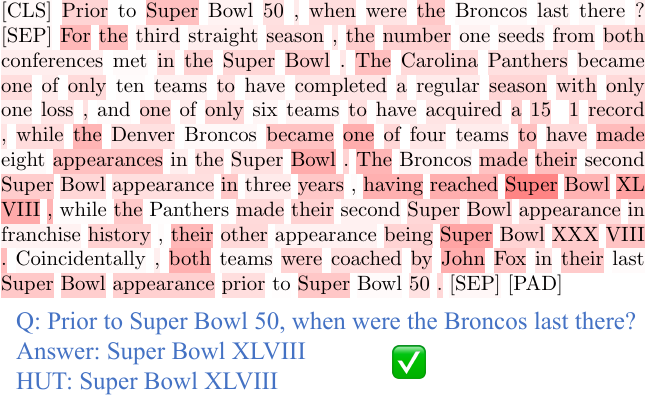}
                \label{HUT_visual}}
\hfill
\subfloat[LoRA]{
                \includegraphics[width=0.47\textwidth]{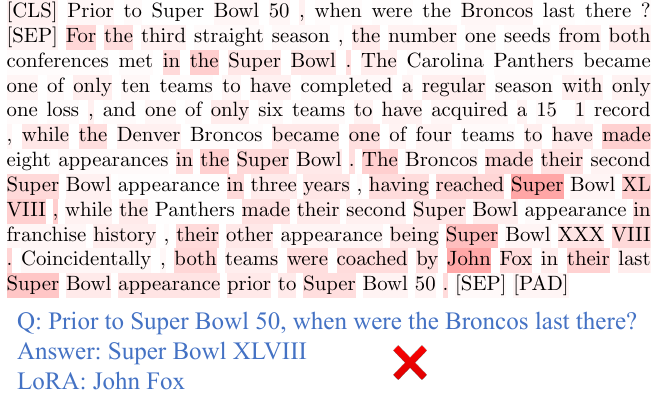}
                \label{lora_visual}}
\vspace{-2mm}
\caption{Visualization of some results. The shades of red indicate the degree of emphasis that the fine-tuned model places on different words. }
\vspace{-5mm}
\label{fig:visualization}
\end{figure*}

\subsubsection{Implementation Details.} We apply HUT to query and value matrices \{$W_q$, $W_v$\} in self-attention module and set $r=4$, using AdamW\cite{DBLP:conf/iclr/LoshchilovH19} optimizer with a linear learning rate schedule for 5 epochs. We keep our setup as close as possible to \cite{DBLP:conf/acl/LiL20} for a direct comparison. The batch size of our methods is set to 4, learning rate is set to 0.002, and beam search beam size is set to 10. Accordingly, we also tune the above hyperparameters for HUT. We report the mean over 3 random seeds; the result for each run is taken from the best epoch. More details please refer to Appendix \ref{sec:experiments}.

\subsubsection{Main Results.} We show in Table \ref{tab:gpt2_ft_results} the experimental results on the E2E NLG challenge after fine-tuning the base model GPT-2 Medium using HUT. As we can see from the table \ref{tab:gpt2_ft_results}, even though our HUT method has far fewer parameters, we are far better than the methods of the Adapter family with more parameters in all five indicators. In addition, compared with the PreLayer and LoRA methods with fewer parameters, our HUT is slightly better than the prelayer method in all metrics, better than the previous SOTA method LoRA in four of five metrics: BLEU, NIST, ROUGE-L and CIDEr, and slightly worse than LoRA in MET. Even compared with the FT method with full parameter update, HUT comprehensively outperforms it. 
Therefore, the above experimental results confirm that HUT can be an effective alternative to full fine-tuning on NLG tasks, and all of them improve the existing PEFT methods to varying degrees. 
So we can confirm that for NLG tasks, our proposed Hadamard updated transformation can also capture the parameter updated features very well with small number of tunable parameters and much lower computation complexity.

\subsection{Ablation Studies}
\subsubsection{Where to apply HUT.} We explored the effect of applying HUT to different attention weight matrices, and the experimental results are shown in Table \ref{tab:weight_type}. We conducted experiments on MRPC and CoLA datasets on the NLU task, and set different $r$ for different subsets of weight matrix in Transformer attention module to ensure consistency of the tunable parameters. From the results, we can find that for the case of applying HUT module to only one weight matrix, applying HUT to $W_k$ matrix has the worst performance while we get the best performance after HUT applying to $W_o$, even outperforming some baseline models. However, When we try to apply HUT to more weight matrices, we find that there is a big drop in performance. We believe that the hadamard transformation in HUT is not suitable for using smaller transformation dimensions $r$ to update all parameter matrices. Instead, it is more suitable to use larger transformation dimensions $r$ to update one or two weight matrix. Therefore, for natural language understanding tasks, to get better performance, HUT should be applied to $W_o$ with large $r$(\textit{e.g.}, 16) or applied to $W_q$ and $W_v$ with a smaller $r$(\textit{e.g.}, 8).

\subsubsection{How to choose $r$.} Another important question is how to choose $r$ to get better performance. We adapt \{$W_q$, $W_v$\}, \{$W_q, W_k, W_v, W_o$\}, and just $W_q$ for a comparison. We evaluate the effect of $r$ on MRPC and CoLA, and results are shown in Table \ref{tab:effect_r}. Surprisingly, we can find that HUT already performs competitively with a very small $r$(\textit{e.g.}, r=1), and this is ture for both just $W_v$ and \{$W_q, W_v$\}. This suggests that the Hadamard updated transformation has a very small "instrisic dimension". As the dimension $r$ goes up, the accuracy for MRPC and the Matthew’s correlation for CoLA does not go up for a meaningful subspace. For \{$W_q, W_k, W_v, W_o$\}, no matter $r$ is small or big, both MRPC and CoLA task have a very bad results, This is the same conclusion as in the previous section.
So there is no need to use  big dimension $r$, HUT can capture enough updated features for weight parameters with a small $r$. This is the reason to ensure the efficiency of computing and  and effectiveness for HUT.

\subsubsection{Visualization.}
To demonstrate the effectiveness of the proposed HUT, we conducted experiments on the SQuADv1.1 \cite{squadv1} dataset.
We visualized the relationship between the output states of the final layer of the fine-tuned model and the inputs in Figure \ref{fig:visualization}, where the varying shades of red indicate the model's attention to that word. From the figure, we can see that the HUT fine-tuned model accurately captures words related to the correct answers and provides the right responses. On the other hand, the LoRA fine-tuned model captures incorrect keywords, leading to wrong answers. Therefore, based on the comparison results, we can infer that the ability of our proposed HUT to capture key features is stronger. We believe this is due to the fact that our proposed UT paradigm can maintain a strong correlation between the learned $\Delta W$ and the original weight $W_0$, and fully leverage the feature encoding capabilities of $W_0$ learned in the pre-training stage.

\section{Conclusion}
In this paper, we propose UT paradigm, which build a direct tranformation to map the original weights to the updated weights. UT paradigm maintain a strong correlation between the pre-trained weight $W_0$ and the updated $\Delta W$. Under this paradigm, we present an approach called HUT. HUT uses Hadamard Transformation which is a powerful feature transformation with only two low-rank matrices to update the original weight matrices. We conduct extensive experiments on NLU and NLG tasks. Results shows that, by using Hadamard transformation, our methods not only achieve on-par or SOTA performance on NLU and NLG tasks, but also reduce the computation complexity during training and inference procedure without introducing any inference latency. Our work demonstrates that the direct updated transformation paradigm of PEFT is feasible.


\section{Limitations}
Despite we proposed a new paradigm for efficiently updating parameters, but we only propose one concrete realization method, called HUT, to verify the effectiveness of this paradigm. Besides, as shown in this paper, our proposed approach has not achieve SOTA on certain datasets, and the exact reason is unknown. So further investigation is necessary to explore the underlying principles of HUT.



\bibliography{acl_latex}

\appendix

\section{Baselines}
\label{sec:baselines}

The details of baseline models are as follows:
\begin{itemize}
    \item \textbf{Full fine-tuning} is the most common approach for adaptation. During fine-tuning, the model is initialized with pre-trained weights and biases, and all model parameters undergo gradient updates.
    \item \textbf{BitFit} \cite{DBLP:conf/acl/ZakenGR22} is an effective parameter-efficient fine-tuning method. The method only fine-tunes bias vectors in the pre-trained model.
    \item \textbf{HAdapter} \cite{DBLP:conf/icml/HoulsbyGJMLGAG19} as proposed to inserts adapter layers between the self-attention module and the MLP module with subsequent residual connection. There are two fully connected layers with biases in an adapter layer with a nonlinearity in between.
    \item \textbf{LAdapter} \cite{DBLP:conf/emnlp/LinMF20} is a more efficient design with the adapter layer applied only after the FFN module and after a LayerNorm \cite{DBLP:journals/corr/BaKH16}.
    \item \textbf{PAdapter} \cite{DBLP:conf/eacl/PfeifferKRCG21} is similiar with LAdapter, adapters only applied after FFN modules and LayerNorm modules \cite{DBLP:journals/corr/BaKH16}.
    \item \textbf{LoRA} \cite{DBLP:conf/iclr/HuSWALWWC22} is most widely used method for PEFT. The number of trainable parameter is controlled by the rank $r$ and the number of adapted weight matrices $n$. We follow the original paper to apply LoRA to query and value projections only.
\end{itemize}

\section{Datasets Details}
\label{sec:datasets}
\subsection{GLUE benchmark} is a wide-ranging collection of natural language understanding tasks. Dataset details are summarized in Appendix It includes MNLI (inference, \cite{DBLP:conf/naacl/WilliamsNB18}), SST-2 (sentiment analysis, \cite{DBLP:conf/emnlp/SocherPWCMNP13}), MRPC (paraphrase detection, \cite{DBLP:conf/acl-iwp/DolanB05}), CoLA (linguistic acceptability, \cite{DBLP:journals/tacl/WarstadtSB19}), QNLI (inference, \cite{DBLP:conf/acl/RajpurkarJL18}), QQP(question-answering), RTE (inference), and STS-B (textual similarity, \cite{DBLP:conf/semeval/CerDALS17}).

\subsection{E2E NLG Challenge} dateset consists of roughly 42,000 training, 4,600 validation, and 4,600 test examples from the restaurant domain. Each source table used as input can have multiple references. Each sample input $(x, y)$ consists of a sequence of slot-value pairs, along with a corresponding natural language reference text. The dataset is released under Creative Commons BY-NC-SA 4.0.

\section{Experiments Details}
\label{sec:experiments}
\subsection{Code Implementation}
We use {\emph{PyTorch}}\footnote{https://pytorch.org/} and {\emph{peft}}\footnote{https://github.com/huggingface/peft} to implement all experiments on NVIDIA RTX 3090 GPUs.
\subsection{Hyperparameters}
For NLU tasks, we train using AdamW with a linear learning rate decay schedule. We sweep learning rate, number of training epochs, and batch size for HUT. we use the hyperparameters presented in Table \ref{tab:hyper_roberta}. 

And for NLG tasks, we train all of our GPT-2 models using AdamW\cite{DBLP:conf/iclr/LoshchilovH19} with a linear learning rate schedule for 5 epochs. We use the batch size, learning rate, and beam search beam size described in\cite{DBLP:conf/acl/LiL20}. Accordingly, we also tune the above hyperparameters for HUT. We report the mean over 3 random seeds. The hyperparameters used for HUT in GPT-2 are listed in Table \ref{tab:hyper_gpt2}.

\begin{table*}[h]
    \footnotesize
    \addtolength{\tabcolsep}{-1pt}
    \centering
    \begin{tabular}{ll|cccccc}
        \hline
        \toprule
        Method  & Dataset     & SST-2 & MRPC & CoLA & QNLI & RTE & STS-B \\
        \midrule
                              & Optimizer   & \multicolumn{6}{c}{AdamW} \\
                              & Warmup Ratio & \multicolumn{6}{c}{0.06} \\
                              & LR Schedule & \multicolumn{6}{c}{Linear} \\
        \midrule
        \multirow{5}{*}{\makecell{RoBERTa large \\ HUT}} \ 
                              & Batch Size & 8 & 16 & 4 & 8 & 16 & 16 \\
                              & \# Epochs & 20 & 20 & 40 & 10 & 80 & 40 \\
                              & Learning Rate & 1E-04 & 5E-03 & 1E-03 & 2E-04 & 2E-03 & 5E-03 \\
                              & HUT Config. & \multicolumn{6}{c}{$r_q=r_v=8$} \\
                              & Max Seq. Len. & \multicolumn{6}{c}{512} \\
        \midrule
        \multirow{5}{*}{\makecell{RoBERTa large \\ LoRA$\dagger$}} \ 
                              & Batch Size & \multicolumn{6}{c}{4} \\
                              & \# Epochs & 10 & 20 & 20 & 10 & 20 & 10 \\
                          & Learning Rate & 4E-04 & 3E-04 & 2E-04 & 2E-04 & 4E-04 & 2E-04 \\
                              & LoRA Config. & \multicolumn{6}{c}{$r_q=r_v=8$} \\
                              & LoRA $\alpha$ & \multicolumn{6}{c}{16} \\
                              & Max Seq. Len. & \multicolumn{6}{c}{128} \\
        \midrule
        \multirow{4}{*}{\makecell{RoBERTa large \\ $\text{Adpt}^\text{P}$ (3M)$\dagger$}} \ 
                              & Batch Size & \multicolumn{6}{c}{32} \\
                              & \# Epochs & 20 & 20 & 20 & 10 & 20 & 20 \\
                              & Learning Rate & 3E-05 & 3E-04 & 3E-04 & 3E-04 & 3E-04 & 3E-04 \\
                              & Bottleneck $r$ & \multicolumn{6}{c}{64} \\
                              & Max Seq. Len. & \multicolumn{6}{c}{128} \\
        \midrule
        \multirow{4}{*}{\makecell{RoBERTa large \\ $\text{Adpt}^\text{P}$ (0.8M)$\dagger$}} \ 
                              & Batch Size & \multicolumn{6}{c}{32} \\
                              & \# Epochs & 20 & 20 & 20 & 10 & 20 & 20 \\
                              & Learning Rate & 3E-04 & 3E-04 & 3E-04 & 3E-04 & 3E-04 & 3E-04 \\
                              & Bottleneck $r$ & \multicolumn{6}{c}{16} \\
                              & Max Seq. Len. & \multicolumn{6}{c}{128} \\
        \midrule
        \multirow{4}{*}{\makecell{RoBERTa large \\ $\text{Adpt}^\text{H}$ (6M)$\dagger$}} \
                              & Batch Size & \multicolumn{6}{c}{32} \\
                              & \# Epochs & 5 & 10 & 10 & 5 & 20 & 10 \\
                              & Learning Rate & 3E-04 & 3E-04 & 3E-04 & 3E-04 & 3E-04 & 3E-04 \\
                              & Bottleneck $r$ & \multicolumn{6}{c}{64} \\
                              & Max Seq. Len. & \multicolumn{6}{c}{128} \\
        \midrule
        \multirow{4}{*}{\makecell{RoBERTa large \\ $\text{Adpt}^\text{H}$ (0.8M)$\dagger$}} \
                              & Batch Size & \multicolumn{6}{c}{32} \\
                              & \# Epochs & 5 & 10 & 10 & 5 & 20 & 10 \\
                              & Learning Rate & 3E-04 & 3E-04 & 3E-04 & 3E-04 & 3E-04 & 3E-04 \\
                              & Bottleneck $r$ & \multicolumn{6}{c}{8} \\
                              & Max Seq. Len. & \multicolumn{6}{c}{128} \\
        \bottomrule
    \end{tabular}
    \caption{The hyperparameters we used for RoBERTa on the GLUE benchmark.}
    \label{tab:hyper_roberta}
\end{table*}

\begin{table*}[h]
\centering
\begin{tabular}{l|c}
\hline
\toprule
Dataset & E2E \\
\midrule
&Training \\
\midrule
Optimizer & AdamW \\
Weight Decay & 0.01 \\
Dropout Prob & 0.1 \\
Batch Size & 4 \\
\# Epoch & 5 \\
Warmup Steps & 500 \\
Learning Rate Schedule & Linear \\
Label Smooth & 0.1  \\
Learning Rate & 0.002 \\
Adaptation & $r_{all}=4$ \\
\midrule
&Inference \\
\midrule
Beam Size & 10 \\
Length Penalty & 0.9  \\
no repeat ngram size & 4 \\
\bottomrule
\end{tabular}
\caption{The hyperparameters for GPT-2 HUT on E2E.}
\label{tab:hyper_gpt2}
\end{table*}

\end{document}